# Evolutionary Algorithm for Graph Coloring Problem

by

**Robiul Islam**

ID: 2009-2-60-004

and

**Arup Kumar Pramanik**

ID: 2009-2-60-008

Bachelors of Science in Computer Science and Engineering

East West University

Supervised By

**Dr. Mozammel Huq Azad Khan**

Professor

Department of Computer Science and Engineering

East West University

**A Project Submitted in Partial Fulfillment of the Requirements for the Degree of Bachelors of Science in Computer Science and Engineering to the**

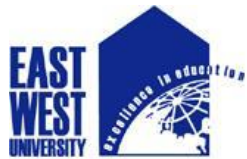

**Department of Computer Science and Engineering**

**East West University**

**Dhaka, Bangladesh**

**April 28, 2013**

# Abstract

The graph coloring problem (GCP) is one of the most studied NP-HARD problems in computer science. Given a graph $G = (V, E)$, the task is to assign a color $c \leq k$ to all vertices $v \epsilon V$ such that no vertices sharing an edge $e \epsilon E$ receive the same color and that the number of used colors, $k$ is minimal. Different heuristic, meta-heuristic, machine learning and hybrid solution methods have been applied to obtain the solution. To solve this problem we use mutation of evolutionary algorithm. For this purpose we introduce binary encoding for Graph Coloring Problem. This binary encoding help us for mutation, evaluate, immune system and merge color easily and also reduce coloring dynamically. In the traditional evolutionary algorithm (EA) for graph coloring, k-coloring approach is used and the EA is run repeatedly until the lowest possible $k$ is reached. In our paper, we start with the theoretical upper bound of chromatic number, that is, maximum out-degree + 1 and in the process of evolution some of the colors are made unused to dynamically reduce the number of color in every generation. We test few standard DIMACS benchmark and compare resent paper. Maximum results are same as expected chromatic color and few data sets are larger than expected chromatic number.

# Declaration

We, hereby declare that, this project was done under CSE-499 thesis Course and has not been submitted elsewhere for the requirements of any degree or diploma or for any other purposes except for publication. This project fulfills the partial requirement for the B.Sc. in Computer Science and Engineering Degree.

\_\_\_\_\_\_\_\_\_\_\_\_\_\_\_\_\_\_

**Robiul Islam**

ID: 2009-2-60-004

\_\_\_\_\_\_\_\_\_\_\_\_\_\_\_\_\_\_\_\_\_\_

**Arup Kumer Pramanik**

ID: 2009-2-60-008

# Letter of Acceptance

I hereby declare that this thesis is from the student's own work and best effort of mine, and all other sources of information used have been acknowledged. This thesis has been submitted with my approval.

___________________________
**Dr. Mozammel Huq Azad Khan**                                    **Supervisor**

Professor

Department of Computer Science and Engineering

East West University

___________________________
**Dr. Nawab Yousuf Ali**                                                     **Chairperson**

Assistant Professor and Chairman

Department of Computer Science and Engineering

East West University

# Acknowledgement

First of all Thanks to **ALLAH** for the uncountable blessings on us.

Thanks to Dr. Mozammel Huq Azad Khan, our supervisor of this thesis works, for his valuable time to help us overcome all the obstacles.

Besides our supervisor, we would like to thank the rest of our thesis committee, and all the members of Computer Science and Engineering faculty for enormous support that they have given us for 4 years.

We again thank our supervisor for the chance to make our thesis about this interesting topic, inspiring talks and the great assistance. Furthermore, we thank all researchers who provided implementations for the Graph Coloring Problem. Without their support, it would not have been possible to carry out this thesis.

Last but not the least; we would like to thank our family for supporting us spiritually throughout our life.

# Abbreviations and Acronyms

EAGCP = Evolutionary Algorithm for Graph Coloring Problem
GCP= Graph Coloring Problem
EA= Evolutionary Algorithm
DIMACS= Discrete Mathematics and Theoretical Computer Science
GA= Genetic Algorithm

# Table of Contents

# List of Figures

# List of Tables

# List of Algorithm

# Chapter 1
# Introduction

In computer science, there are some problems which arise more frequently and consequently, well-investigated. One of these is the prominent graph colouring problem (GCP), which is one of Karp's NP-COMPLETE problems [2]. This problem has its origins in colouring the countries map such that no neighbouring countries receive the same colour. In this context, a colouring of a graph $G = (V, E)$ is an assignment of a colour $c \leq k$ to all vertices $v \in V$ of the graph such that no adjacent vertices $u, v \in V : (u, v) \in E$ receive the same colour and that the number of used colours. Although this sounds easy, finding such a colouring with only a limited number of colours can be very hard. Even more, due to its NP-completeness, it is unlikely (unless P = NP) that there exist exact strategies which require less than exponential time to colour an arbitrary graph. As a result, much focus has been spent on the development of (Meta) heuristics approaches for the problem. This method does not ensure optimal solutions, but return good colouring in a reasonable time. For the GCP, various algorithms have been developed, starting with greedy algorithms [2] to more sophisticated techniques from the area of (Meta) heuristics like local search or genetic algorithms. Some of the most popular solvers in this context base on a Tabu Search [2], Variable Neighbourhood [2] or Iterated Local Search [2]. Other methods built on Genetic Algorithms [2] (GA )or Ant Colony Optimization [2],Map Colouring [5], Scheduling [6, 17], Radio Frequency Assignment [7, 17], Register allocation [8, 17], Pattern Matching [1], Sudoku [1], Wisdom of the Crowds [1].

## 1.1 Objective
The objectives of the thesis are;
    I.    Confirm minimum colour of tested graph of this paper of standard DIMACS benchmark
    II.    Compares our thesis result with other researcher paper

## 1.2 Main Result
The main result of this thesis is maximum dataset are match to our result

Graph colouring problem is so easy to describe, but difficult to solve. This problem belongs to NP-hard [2] problems. Colouring such a way that will satisfy all constrains of the problem and find the optimal solution is a hard part but here is our algorithm which works almost perfectly to solve the problems.

In this paper we present Evolutionary Algorithm for Graph Colouring Problem (EAGCP) to solve graph colouring problem. This algorithm can treat balance between exploration and exploitation more easily compared to other algorithms. The result shows that the algorithm gains remarkable experimental results and some solutions are optimal and some are near optimal.

# Chapter 2
# Background

## 2.1 Graph Colouring Problem

A colouring of simple graph is the assignment of a colour to each vertex of the graph so that no two adjacent vertices are assigned the same colour.

Let G be an undirected graph with no loops. A *k*-colouring of G is an assignment of *k* colours to the vertices of G in such a way that adjacent vertices are assigned different colours. If G has a *k*-colouring, then G is said to be *k*-colourable. The chromatic number of G, denoted by $\chi(G)$, is the smallest number *k* for which is *k*-colourable. For example,

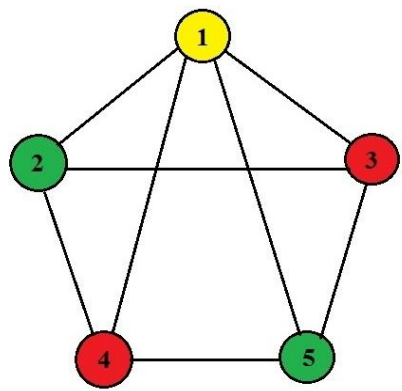

Figure 2.1: A graph having colourful vertex

The first results about graph colouring deal almost exclusively with planar graphs in the form of the colouring of maps. While trying to colour a map of the counties of England, Francis Guthrie postulated the four colour conjecture, noting that four colours were sufficient to colour the map so that no regions sharing a common border received the same colour. Guthrie's brother passed on the question to his mathematics teacher Augustus de Morgan at University College, who mentioned it in a letter to William Hamilton in 1852. Arthur Cayley raised the problem at a meeting of the London Mathematical Society in 1879. The same year, Alfred Kempe published a paper that claimed to establish the result, and for a decade the four colour problem was considered solved. For his accomplishment Kempe was elected a Fellow of the Royal Society and later President of the London Mathematical Society [20]. The chromatic number of a graph is the least number of colours needed for a colouring of this graph.

In 1890, Heawood pointed out that Kempe's argument was wrong. However, in that paper he proved the five colour theorem, saying that every planar map can be coloured with no more than five colours, using ideas of Kempe. In the following century, a vast amount of work and theories were developed to reduce the number of colours to four, until the four colour theorem was finally proved in 1976 by Kenneth Appel and Wolfgang Haken. Perhaps surprisingly, the proof went back to the ideas of Heawood and Kempe and largely disregarded the intervening developments.[21] The proof of the four colour theorem is also noteworthy for being the first major computer-aided proof.

In 1912, George David Birkhoff introduced the chromatic polynomial to study the colouring problems, which was generalised to the Tutte polynomial by Tutte, important structures in algebraic graph theory. Kempe had already drawn attention to the general, non-planar case in 1879,[22] and many results on generalisations of planar graph colouring to surfaces of higher order followed in the early 20th century.

In 1960, Claude Berge formulated another conjecture about graph colouring, the strong perfect graph conjecture, originally motivated by an information-theoretic concept called the zero-error capacity of a graph introduced by Shannon. The conjecture remained unresolved for 40 years, until it was established as the celebrated strong perfect graph theorem in 2002 by Chudnovsky, Robertson, Seymour, Thomas 2002.

Graph colouring has been studied as an algorithmic problem since the early 1970s: the chromatic number problem is one of Karp's 21 NP-complete problems from 1972, and at approximately the same time various exponential-time algorithms were developed based on backtracking and on the deletion-contraction recurrence of Zykov (1949). One of the major applications of graph colouring, register allocation in compilers was introduced in 1981.

## 2.2 Evolutionary Algorithm

Evolutionary algorithms (EA) are search algorithm based on the mechanics of natural selection and natural genetics. They combine survival of the fittest among string structures with a structured yet randomized information exchange to form a search algorithm with some of the innovative flair of human search. In every generation, a new set of artificial creatures (strings) is crested using bits and pieces of the fittest of the old; an occasional new part is tried for good measure [23]. Overall search technique are given bellow

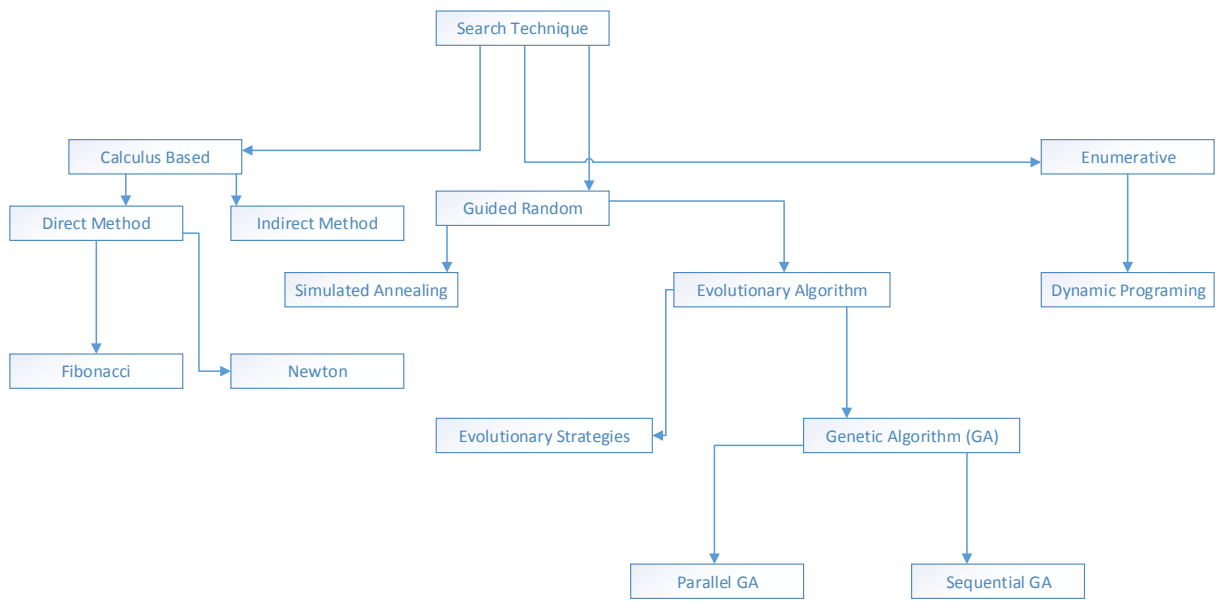

Figure 2.2: Classification of search technique based on [24]

Evolutionary algorithm is a process which improves the problem result output generation by generation having treat balance between exploration and exploitation more easily. An EA uses mechanisms inspired by biological evolution, such as reproduction, mutation, recombination, and selection. One of the possible limitations of many evolutionary algorithms is their lack of a clear genotype-phenotype distinction. But here in our algorithm we use binary encoding technique to solve this problem. An evolutionary algorithm is essentially a type of genetic algorithm in which the virtual chromosomes are made of real values instead of some form of bit representation.

The idea of evolutionary computing was introduced in 1960 by I. Rechenberg in his work Evolutionary strategies. Genetic Algorithm (GA) is computerized search and optimization algorithms based on the mechanics of natural genetics and natural selection. Prof. Holland of university of Michigan, Ann Arbor, envisaged the concept of these algorithms in the mid-sixties and published his seminal work (Holland, 1975). Thereafter, a number of student and other researcher have contributed to the development of the field.

To date, most of the GA studies are available through some books by Davis (1991), Goldberg (1989), Holland (1975), MIchalewicz (1992) and Deb (1995) and through a number of conference proceeding. The first application toward structural engineering was carried by Goldberg and Samtani (1986). They applied GA to the optimization of a ten-member plane truss. Jenkins (1991) applied GA to a trussed beam structure. Deb (1991) and Rajeev and Krishnamoorthy (1992) has also applied GA to structural engineering problem. Apart from structural engineering there are many other fields in which GA's have been applied successfully. It includes biology, computer science, image processing and pattern recognition, physical science, social science and neural network.

## 2.3 Mutation

Mutation is the process to change or damage the DNA sequence consequently and used to maintain genetic diversity from one generation of a population of genetic algorithmchromosomesto the next generation. Mutation is an important part of the genetic algorithm as helps to prevent the population from stagnating at any local optima.In this process we need some better chromosome or damage chromosome. Anything is happened based on Adenine (A), Guanine (G), Cytosine (C), and Thymine (T) [17].

Genetic Server and Genetic Library include the following types of mutation:

**Flip Bit** -A mutation operator that simply inverts the value of the chosen gene (0 goes to 1 and 1 goes to 0). This mutation operator can only be used for binary genes.

**Boundary** - A mutation operator that replaces the value of the chosen gene with either the upper or lower bound for that gene (chosen randomly). This mutation operator can only be used for integer and float genes.

**Non-Uniform** - A mutation operator that increases the probability that the amount of the mutation will be close to 0 as the generation number increases. This mutation operator keeps the population from stagnating in the early stages of the evolution then allows the genetic algorithm to fine tune the solution in the later stages of evolution. This mutation operator can only be used for integer and float genes.

**Uniform** - A mutation operator that replaces the value of the chosen gene with a uniform random value selected between the user-specified upper and lower bounds for that gene. This mutation operator can only be used for integer and float genes.

**Gaussian** - A mutation operator that adds a unit Gaussian distributed random value to the chosen gene. The new gene value is clipped if it falls outside of the user-specified lower or upper bounds for that gene. This mutation operator can only be used for integer and float genes.

## 2.4 DIMACS

The Centre for Discrete Mathematics and Theoretical Computer Science (DIMACS) is collaboration between Rutgers University, Princeton University, and the research firms AT&T, Bell Labs, telcordia, and NEC. It was founded in 1989 with money from the National Science Foundation. Its offices are located on the Rutgers campus, and 250 members from the six institutions form its permanent members.

DIMACS is devoted to both theoretical development and practical applications of discrete mathematics and theoretical computer science. It engages in a wide variety of evangelism including encouraging, inspiring, and facilitating researchers in these subject areas, and sponsoring conferences and workshops.

# Chapter 3

# Related Work

We know in computer science Graph Colouring problem is most prominent and it uses in different situation to solve different problem. The convention of using colours originates from colouring the countries of a map, where each face is literally collared. This was generalized to colour the faces of a graph embedded in the plane. Actually it became colouring the vertices, and in this form it generalizes to all graphs. In computer representations, it is typical to use the first few positive integers as the "colours". In general, one can use any finite set as the "colour set". The nature of the colouring problem depends on the number of colours but not on what they are Graph colouring enjoys many practical applications as well as theoretical challenges. Beside the classical types of problems, different limitations can also be set on the graph, or on the way a colour is assigned, or even on the colour itself. It is still a very active field of research.

## 3.1: Multipoint Guided Mutation for GCP [3]

This process can reduce colour randomly performing crossover and mutation based on their individual chromosome. In multipoint special mutation part, it can reduce chromatic number and cheek where chromosome is valid or not accounting to their fitness value.

## 3.2: Wisdom of Artificial Crowds [1]

This chromosome can reduce colour in best half of final population based on best performing chromosome. Here algorithm priority most used colour. With based on most used colour algorithm influence other chromosome to used minimum chromatic number.

## 3.3: Hierarchical Parallel GA for GCP [25]

This process can create a list of node based on their degrees in a decreasing order. Then insert one node and apply his algorithm for inserting sub-graph. This process run until every node has given colour.

## 3.4 Greedy Colouring

The greedy algorithm considers the vertices in a specific order $vi \ldots \ldots \ldots vn$ and assigns to the smallest available colour not used by $vi's$ neighbours among $vi \ldots \ldots v(i-1)$ adding a fresh colour if needed. The quality of the resulting colouring depends on the chosen ordering. There exists an ordering that leads to a greedy colouring with the optimal number of $\chi(g)$ colours.

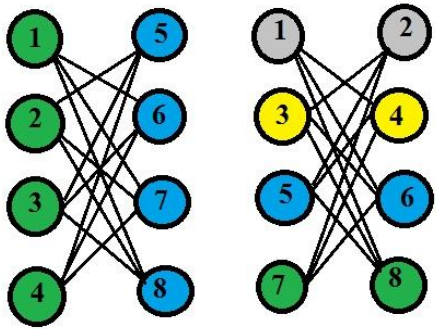

Figure 3.1: Two greedy colouring of the same graph using different vertex order

The right example generalises to 2-colorable graphs with $n$ vertices, where the greedy algorithm expends $n/2$ colors.

## 3.5: Parallel and Distributed Algorithms for GCP

A parallel or distributed algorithm is executed by a number of processes and/or processors to accomplish a particular task. Generally, it is much harder to design correct parallel protocols than their sequential counterparts. The reason is that it is hard to imagine all possible behaviours of a parallel system. The terms "concurrent computing", "parallel computing", and "distributed computing" have a lot of common, and no clear distinction exists between them. The problem of edge colouring has also been studied in the distributed model. The lower bound for distributed vertex colouring due to Linial (1992) applies to the distributed edge colouring problem as well.

# Chapter 4
# Proposed Evolutionary Algorithm for Graph Colouring Problem (EAGCP)

Basic Algorithm of Evolutionary Algorithm for Graph Colouring Problem (EAGCP) is below

    Algorithm: EAGCP

    Population initialization randomly

    Loop until terminated

        Fitness Calculation

        Population correction

        Checking duplication

        Copying total population in temporary population

        Mutation in temporary population

        Repairing temporary population

        Fitness calculation to temporary

        Temporary population correction

        Marge temporary population to initial population according to minimum fitness that no duplicate chromosome cannot be allow

        Adding new population

        If minimum fitness is equal to total number of population

            Call deterministic process

            Calculate fitness

                If new minimum fitness is equal to old minimum fitness

                    Terminate

                Else

                    Replace old population to new population

            End

        End deterministic

    End loop

    End Algorithm

Algorithm 4.1: Proposed Evolutionary Algorithm for Graph Colouring Problem (EAGCP)

Here, we use binary encoding for this problem. We start for theoretical upper bound that is the maximum out degree +1 in row and number of node is in column. We take the example myciel3.col from international stranded Graph Colouring Instances website [4] DIMACS where node, n =11 edge, e= 20 and maximum out degree, m =5. So we start a binary matrix contains colour row m+1 and column n.

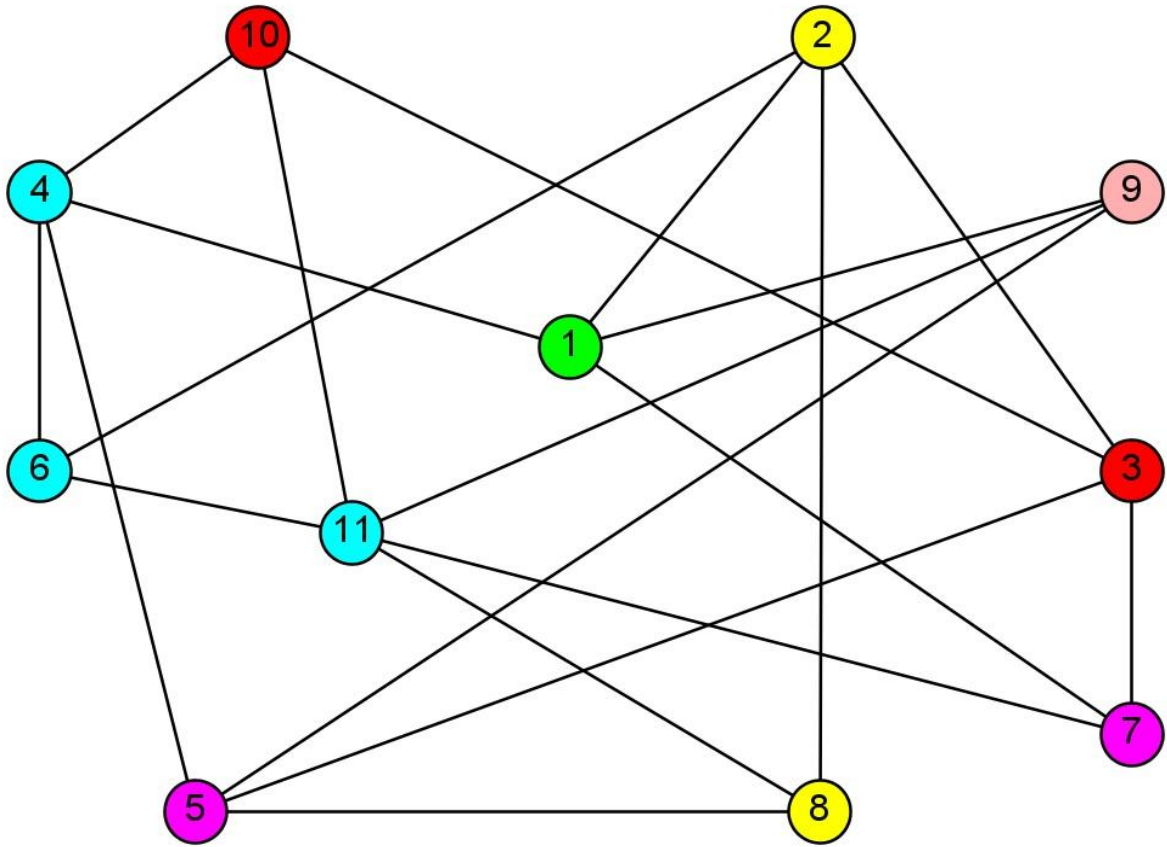

Figure 4.1: Graph of myciel3.col from standard DIMACS benchmark

## 4.1 Population Initialization

From this above graph we create a randomly binary matrix with Colour *Node. In figure 4.1 Colour is maximum out degree +1 which is represent row of binary matrix and nodes of the graph represent column of this matrix. In this data set node 11 has maximum out degree 5. So we create a binary matrix having (5+1)*11. Initially fill the entire index randomly 0 or 1. Example of a single population are given bellow

| 0 | 0 | 1 | 0 | 0 | 0 | 0 | 0 | 0 | 1 | 0 |
|---|---|---|---|---|---|---|---|---|---|---|
| 0 | 0 | 0 | 0 | 1 | 0 | 1 | 0 | 0 | 0 | 0 |
| 0 | 1 | 0 | 0 | 0 | 0 | 0 | 1 | 0 | 0 | 0 |
| 0 | 0 | 0 | 1 | 0 | 1 | 0 | 0 | 0 | 0 | 1 |
| 0 | 0 | 0 | 0 | 0 | 0 | 0 | 0 | 1 | 0 | 0 |
| 1 | 0 | 0 | 0 | 0 | 0 | 0 | 0 | 0 | 0 | 0 |

Table 4.1: Initial matrix for myciel3.col data set

## 4.2 Repairing Population

This process ensures that every column have single one because we know that one node can contain only one colour. Here in this chromosome matrix we replace single zero to one for every column randomly.

Repairing process works when the number of one in a single column is more than one then it will pick a random position make it one and all other one will convert into zero whether it one or zero. When more than one is found in any column, it's assuming that one node has more than one colour which is impossible. That's why it is needed to keep single 1 in every column. This way repairing process works. Here is a sample example to repair a chromosome.

| 0 | 0 | 1 | 0 | 0 | 1 | 0 | 0 | 0 | 1 | 0 |
|---|---|---|---|---|---|---|---|---|---|---|
| 0 | 0 | 0 | 1 | 1 | 0 | 0 | 0 | 0 | 0 | 1 |
| 0 | 1 | 0 | 0 | 0 | 0 | 0 | 1 | 0 | 0 | 1 |
| 0 | 0 | 0 | 1 | 0 | 1 | 0 | 0 | 0 | 0 | 1 |
| 0 | 0 | 1 | 0 | 0 | 0 | 0 | 1 | 1 | 0 | 0 |
| 0 | 1 | 1 | 1 | 0 | 0 | 0 | 0 | 0 | 0 | 0 |

Table 4.2: Initial chromosome of data set

In table 4.2 shows that node 1 has no colour, node two has 2 colours which violates the rule of graph colouring problem (GCP). This function ensures that every node contains one single colour. Here is the previous example after this process

| 0 | 0 | 1 | 0 | 0 | 0 | 0 | 0 | 0 | 1 | 0 |
|---|---|---|---|---|---|---|---|---|---|---|
| 0 | 0 | 0 | 0 | 1 | 0 | 1 | 0 | 0 | 0 | 0 |
| 0 | 1 | 0 | 0 | 0 | 0 | 0 | 1 | 0 | 0 | 0 |
| 0 | 0 | 0 | 1 | 0 | 1 | 0 | 0 | 0 | 0 | 1 |
| 0 | 0 | 0 | 0 | 0 | 0 | 0 | 0 | 1 | 0 | 0 |
| 1 | 0 | 0 | 0 | 0 | 0 | 0 | 0 | 0 | 0 | 0 |

Table 4.3: Correctchromosome after repairing

So the repairing is that process which can ensure every column should have single 1 that actually confirms one node gets one colour. So when there is not found any 1 in a single column, pick a position randomly and replace it by 1.

## 4.3 Fitness Calculation

For calculating the fitness of a single chromosome we introduce a penalty function [15]. Here in our algorithm, we take penalty = m+1 which is its initial colour. The main benefit of this function is taking the minimum fitness value and discards the maximum value. After any mutation, if the fitness value is smaller than the previous value then replaces the bigger one to the smaller one. The equation mathematically looks like

$$Fitness\ Function, f(x) = Valid\ Colour + Invalid\ Colour * penalty \qquad (eq\ 4.1)$$

To confirm valid and invalid colour use [4] for this. Here for simplicity we use adjacency matrix of that graph. Here, adjacency graph of [4] myciel3.col

| node/node | 1 | 2 | 3 | 4 | 5 | 6 | 7 | 8 | 9 | 10 | 11 |
|---|---|---|---|---|---|---|---|---|---|---|---|
| 1 | 0 | 1 | 0 | 1 | 0 | 0 | 1 | 0 | 1 | 0 | 0 |
| 2 | 1 | 0 | 1 | 0 | 0 | 1 | 0 | 1 | 0 | 0 | 0 |
| 3 | 0 | 1 | 0 | 0 | 1 | 0 | 1 | 0 | 0 | 1 | 0 |
| 4 | 1 | 0 | 0 | 0 | 1 | 1 | 0 | 0 | 0 | 1 | 0 |
| 5 | 0 | 0 | 1 | 1 | 0 | 0 | 0 | 1 | 1 | 0 | 0 |
| 6 | 0 | 1 | 0 | 1 | 0 | 0 | 0 | 0 | 0 | 0 | 1 |
| 7 | 1 | 0 | 1 | 0 | 0 | 0 | 0 | 0 | 0 | 0 | 1 |
| 8 | 0 | 1 | 0 | 0 | 1 | 0 | 0 | 0 | 0 | 0 | 1 |
| 9 | 1 | 0 | 0 | 0 | 1 | 0 | 0 | 0 | 0 | 0 | 1 |
| 10 | 0 | 0 | 1 | 1 | 0 | 0 | 0 | 0 | 0 | 0 | 1 |
| 11 | 0 | 0 | 0 | 0 | 0 | 1 | 1 | 1 | 1 | 1 | 0 |

Table 4.4: Adjacency matrix of myciel3.col data file base on [4]

First we know that how we confirm valid colour or invalid colour.

Rules of valid colour

      i.     If one row has one 1 that indicate one colour in this particular node.

     ii.    When a single row has more than one 1 and there is no edge [2, 4] between those two edges that is also valid row and colour is also valid. Therefore one

colour goes to one or two or many node that does not have any edge between them [2].

Rules of Invalid colour

i. In this algorithm total colour = maximum out degree +1 which we know from our theoretical background.

ii. Unused colour means in that the row does not contain any one. Therefore this colour cannot represent any node of this chromosome.

Or we have another option to identify in invalid colour that is one colour goes two or more than two edges which have share edges.

$Invalid\ Colour =$
$Total\ Colour - (Unused\ Colour + Valid\ Colour)$  (eq 4.2)

How we calculate fitness function of every chromosome are given bellow

| 0 | 0 | 1 | 0 | 0 | 1 | 0 | 0 | 0 | 0 | 1 |
|---|---|---|---|---|---|---|---|---|---|---|
| 0 | 0 | 0 | 0 | 0 | 0 | 1 | 0 | 0 | 0 | 0 |
| 0 | 0 | 0 | 0 | 0 | 0 | 0 | 0 | 0 | 0 | 0 |
| 0 | 0 | 0 | 0 | 1 | 0 | 0 | 1 | 0 | 0 | 0 |
| 1 | 0 | 0 | 0 | 0 | 0 | 0 | 0 | 0 | 0 | 0 |
| 0 | 1 | 0 | 1 | 0 | 0 | 0 | 0 | 1 | 1 | 0 |

Table 4.5: Invalid Chromosome of myciel3.col

Here, Invalid Colour, $I_c$ = 3

Valid Colour, $V_c$ = 2

Unused Colour, $U_c$ = 1

Penalty, P = 6

| $I_c$ | 0 | 0 | 1 | 0 | 0 | 1 | 0 | 0 | 0 | 0 | 1 |
|---|---|---|---|---|---|---|---|---|---|---|---|
| $V_c$ | 0 | 0 | 0 | 0 | 0 | 0 | 1 | 0 | 0 | 0 | 0 |
| $U_c$ | 0 | 0 | 0 | 0 | 0 | 0 | 0 | 0 | 0 | 0 | 0 |
| $I_c$ | 0 | 0 | 0 | 0 | 1 | 0 | 0 | 1 | 0 | 0 | 0 |
| $V_c$ | 1 | 0 | 0 | 0 | 0 | 0 | 0 | 0 | 0 | 0 | 0 |
| $I_c$ | 0 | 1 | 0 | 1 | 0 | 0 | 0 | 0 | 1 | 1 | 0 |

Table 4.6: Details of table 4.5

So, fitness value of this chromosome is based on $eq\ 4.1$ is $f(x) = 2 + 3 * 6 = 20$.

In 3${}^{rd}$ row no of one is found zero or this colour cannot be assign to any node of this specific graph. In case of invalid colour here 1${}^{st}$colour node 1-6 has 1${}^{st}$colour but there is a shared edge between 1 and 6. So this colour is invalid. In our case valid colour shared edge does not have same colour. In this process invalid edge multiply by penalty [15] function increase the whole fitness value. So minimum number of colour used will be consider as best chromosome and maximum number chromosome is eliminated by iteration process of generic algorithm. We introduce more than one chromosome depend on difference data set of DIMACS [4] and kept lower colour chromosome for the next iteration.

## 4.4 Population Correction

To correct our population those rows are invalid and move that invalid one randomly upper or lower position in that column so that that row becomes valid row. For the continuation of this process we select those populations whose number of chromosome is gather than m+1. When fitness value is less than m+1, all colours are valid and that indicates that the no of invalid colour become zero. Here is an example that contains three invalid colours in three nodes

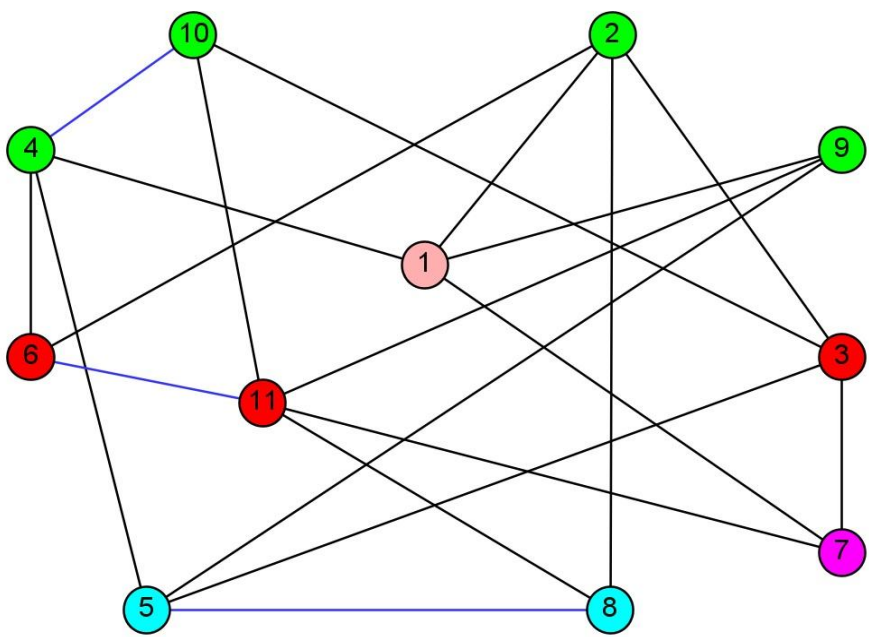

Figure 4.2: Invalid chromosome

First we have to define which node colour is invalid. In this chromosome, node 6 and node 11, node 4 and node 10, node 5 and node 8 have shared edge with same colour. So we have to change nodes colour randomly, so that 1st colour row is valid

| V$_c$ | 0 | 0 | 1 | 0 | 0 | 0 | 0 | 0 | 0 | 0 | 1 |
| V$_c$ | 0 | 0 | 0 | 0 | 0 | 0 | 1 | 0 | 0 | 0 | 0 |
| V$_c$ | 0 | 0 | 0 | 0 | 0 | 1 | 0 | 0 | 0 | 0 | 0 |
| I$_c$ | 0 | 0 | 0 | 0 | 1 | 0 | 0 | 1 | 0 | 0 | 0 |
| V$_c$ | 1 | 0 | 0 | 0 | 0 | 0 | 0 | 0 | 0 | 0 | 0 |
| I$_c$ | 0 | 1 | 0 | 1 | 0 | 0 | 0 | 0 | 1 | 1 | 0 |

Table 4.7: Changing colour of Node 6

Fitness Function of Table 4.8 based on 4.5, $4 + 2 * 6 = 16$

Then find another invalid colour and doing the same process. Now 4$^{th}$ colour row is invalid and changing that position randomly with 3$^{rd}$ colour row

| V$_c$ | 0 | 0 | 1 | 0 | 0 | 0 | 0 | 0 | 0 | 0 | 1 |
| V$_c$ | 0 | 0 | 0 | 0 | 0 | 0 | 1 | 0 | 0 | 0 | 0 |
| V$_c$ | 0 | 0 | 0 | 0 | 1 | 1 | 0 | 0 | 0 | 0 | 0 |
| V$_c$ | 0 | 0 | 0 | 0 | 0 | 0 | 0 | 1 | 0 | 0 | 0 |
| V$_c$ | 1 | 0 | 0 | 0 | 0 | 0 | 0 | 0 | 0 | 0 | 0 |
| I$_c$ | 0 | 1 | 0 | 1 | 0 | 0 | 0 | 0 | 1 | 1 | 0 |

Table 4.8: Changing colour of Node 5

Fitness function of Table 4.9 based on 4.5, $5 + 1 * 6 = 11$

Here a point can be noted that colour 3$^{rd}$ colour row contains more than 1. But it is valid because those two nodes don't have a common shared edge. Now find another invalid colour and do it as same things. Now 6$^{th}$ colour row is invalid and move that position randomly. Here is 4$^{th}$ colour row

| V$_c$ | 0 | 0 | 1 | 0 | 0 | 0 | 0 | 0 | 0 | 0 | 1 |
| V$_c$ | 0 | 0 | 0 | 0 | 0 | 0 | 1 | 0 | 0 | 0 | 0 |
| V$_c$ | 0 | 0 | 0 | 0 | 1 | 1 | 0 | 0 | 0 | 0 | 0 |
| V$_c$ | 0 | 0 | 0 | 1 | 0 | 0 | 0 | 1 | 0 | 0 | 0 |
| V$_c$ | 1 | 0 | 0 | 0 | 0 | 0 | 0 | 0 | 0 | 0 | 0 |
| V$_c$ | 0 | 1 | 0 | 0 | 0 | 0 | 0 | 0 | 1 | 1 | 0 |

Table 4.9: Changing colour of Node 4

Fitness function of Table 4.7, $6 + 0 * 6 = 6$

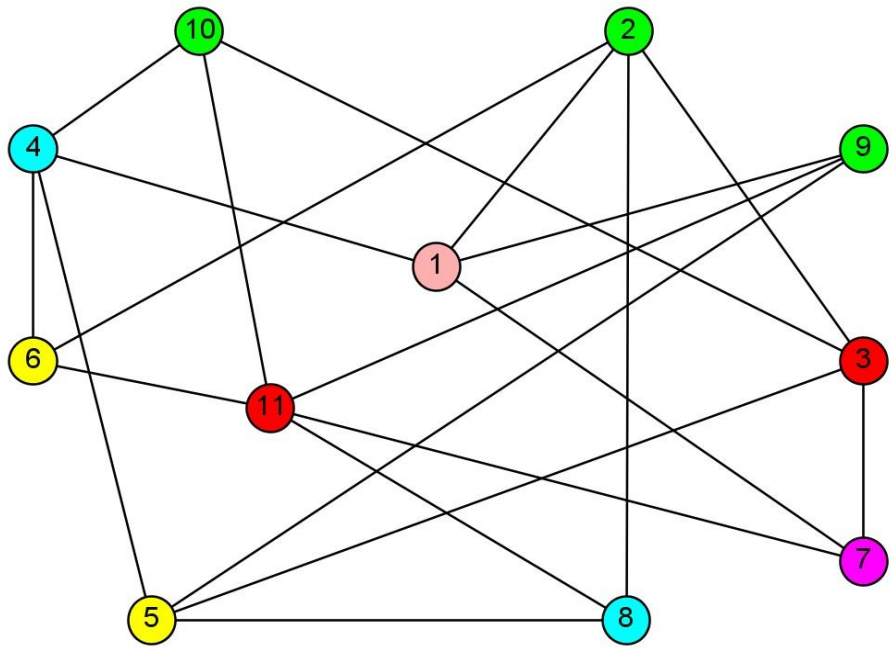

Figure 4.3: Valid chromosome

Now all colours are valid after finish reparation. There is no chromosome who contains invalid colour. We keep this chromosome for another part of our algorithm. Particularly we correct all chromosomes in every population. This makes iteration time faster and quickly. So that finding the minimum number of colour becomes easier.

### 4.5 Checking Duplication

For checking duplication in a single chunk of chromosome no duplicate chromosome is allowed. The chromosome which is repeated one or more times that is a huge problem which may causes local optima. So we have to replace that chromosome by a new randomly corrected chromosome which is also non duplicate. Here chunk means number of population which varies difference dataset of [4].

### 4.6 Copying Total Population to Temporary Population

In this section we create another population which size is equals to the original population sizeand copy the original population into it.

## 4.7 Mutation

In our research work we use actually flip bit type mutation technique. We fully mutated the population in this process. This happens based on some probability function. The mutation process runs when the mutation probability is less then 0.1 and then flips a bit randomly. Mutation probability depends on difference data file of DIMACS graph. In this process we select an index randomly with low probability and change that bit permanently. If probability is set to high, finding the optimal solution becomes harder. This process will go to the entire chromosome throughout the total population. We improve all of chromosome one and only based on this mutation. How process work is are given bellow:

| 0 | 0 | 1 | 0 | 0 | 0 | 0 | 0 | 0 | 0 | 1 |
|---|---|---|---|---|---|---|---|---|---|---|
| 0 | 0 | 0 | 0 | 0 | 0 | 1 | 0 | 0 | 0 | 0 |
| 0 | 0 | 0 | 0 | 1 | 1 | 0 | 0 | 0 | 0 | 0 |
| 0 | 0 | 0 | 1 | 0 | 0 | 0 | 1 | 0 | 0 | 0 |
| 1 | 0 | 0 | 0 | 0 | 0 | 0 | 0 | 0 | 0 | 0 |
| 0 | 1 | 0 | 0 | 0 | 0 | 0 | 0 | 1 | 1 | 0 |

Table 4.10: Initial population

Now we can create another two matrixes randomly. First one for the Probability matrix which is 0 to 1 range and the other one is for the mutation position which bit we want to change. Its range is 1 to total node number. Here is an example for the data set myciel3.col

| (a) | (b) |
|---|---|
| 0.10571 | 11 |
| 0.14204 | 9 |
| 0.06646 | 7 |
| 0.42096 | 1 |
| 0.57371 | 8 |
| 0.05208 | 2 |

Table 4.11: (a) is probability between 0 to 1 and (b) consequent random colour node

Suppose here mutation probability to 10%. So when value is less than 0.1 in (a), (b) number of node that position will change. Here the table (a) $3^{rd}$ and $6^{th}$ row value is less than 0.1 and in the table (b) random number generator will generate six random numbers between 1 to total node. Here change the chromosomes $3^{rd}$ rows $7^{th}$ position and $6^{th}$ rows $2^{nd}$ number position

| 0 | 0 | 1 | 0 | 0 | 0 | 0 | 0 | 0 | 0 | 1 |
|---|---|---|---|---|---|---|---|---|---|---|
| 0 | 0 | 0 | 0 | 0 | 0 | 1 | 0 | 0 | 0 | 0 |
| 0 | 0 | 0 | 0 | 1 | 1 | 1 | 0 | 0 | 0 | 0 |
| 0 | 0 | 0 | 1 | 0 | 0 | 0 | 1 | 0 | 0 | 0 |
| 1 | 0 | 0 | 0 | 0 | 0 | 0 | 0 | 0 | 0 | 0 |
| 0 | 0 | 0 | 0 | 0 | 0 | 0 | 0 | 1 | 1 | 0 |

Table 4.12: Sample Chromosome after mutation

In our algorithm we have changed mutation probability different time to check, in which probability the optimal output comes faster and efficiently.

## 4.8 Merging Temporary Population to Initial Population

Fitness value of both populations are given bellow

| Fitness Value of Chromosome | 4 | 5 | 4 | 6 | 4 | 5 | 4 | 5 | 4 |
|---|---|---|---|---|---|---|---|---|---|
| Fitness Value of Temporary Chromosome | 4 | 6 | 5 | 6 | 5 | 5 | 4 | 5 | 6 |

Table 4.13: Fitness value of original and temporary population

In this process we first target maximum value of main fitness and minimum value of temporary fitness. Then we move the minimum fitness to the main population. Process is show in bellow

| Fitness Value of Chromosome | 4 | 5 | 4 | 4 | 4 | 5 | 4 | 5 | 4 |
|---|---|---|---|---|---|---|---|---|---|
| Fitness Value of Temporary Chromosome | 6 | 6 | 5 | 6 | 5 | 5 | 4 | 5 | 6 |

Table 4.14: Replacing minimum fitness step 1

| Fitness Value of Chromosome | 4 | 5 | 4 | 4 | 4 | 4 | 4 | 5 | 4 |
|---|---|---|---|---|---|---|---|---|---|
| Fitness Value of Temporary Chromosome | 6 | 6 | 5 | 6 | 5 | 5 | 5 | 5 | 6 |

Table 4.15: Replacing minimum fitness step 2

Now maximum value of main fitness is 5 and minimum value of temporary fitness is also 5. So we stop here. In this process we replace some bad chromosome to good ones.

## 4.9 Adding New Population

Now we add some presence of population base on database [4]. This population is adding to replace worst population of main chromosome. For example in this chromosome here we have 9 chromosomes. Suppose, we adding 20% new population in this process. Here when we round it up we found 2 correct populations. So we choose worst 2 populations and replace them with a new random population without considering the fitness value so that we can avoid the local optima problem. For example

| Fitness Value of Chromosome | 4 | 5 | 4 | 4 | 4 | 4 | 4 | 5 | 4 |
| --- | --- | --- | --- | --- | --- | --- | --- | --- | --- |
| Fitness Value of Chromosome after adding | 4 | 4 | 4 | 4 | 4 | 4 | 4 | 6 | 4 |

Table 4.16: Adding new population

For adding new population of evolutionary process it can auto solve problem of local optima. In biology this process is called Immune system. In biological background this Immune process helps to protects against disease.

## 4.10 Deterministic Process

This process cans marge colour correctly. Less used colour can be merged by the more used colour. In this process $1^{st}$ nodes colour is replaced by $11^{th}$ nodes colour because pink colour is less uses then red and $7^{th}$ nodes colour is replaced by $5^{th}$ nodes colour for same reason. Finding which colour is less use then replace it by the more used colour makes sure that the chromosome contains all valid colours. So in this way it can minimize the total valid colour. If the fitness value does remain same in a certain number of iteration then our algorithm stops.

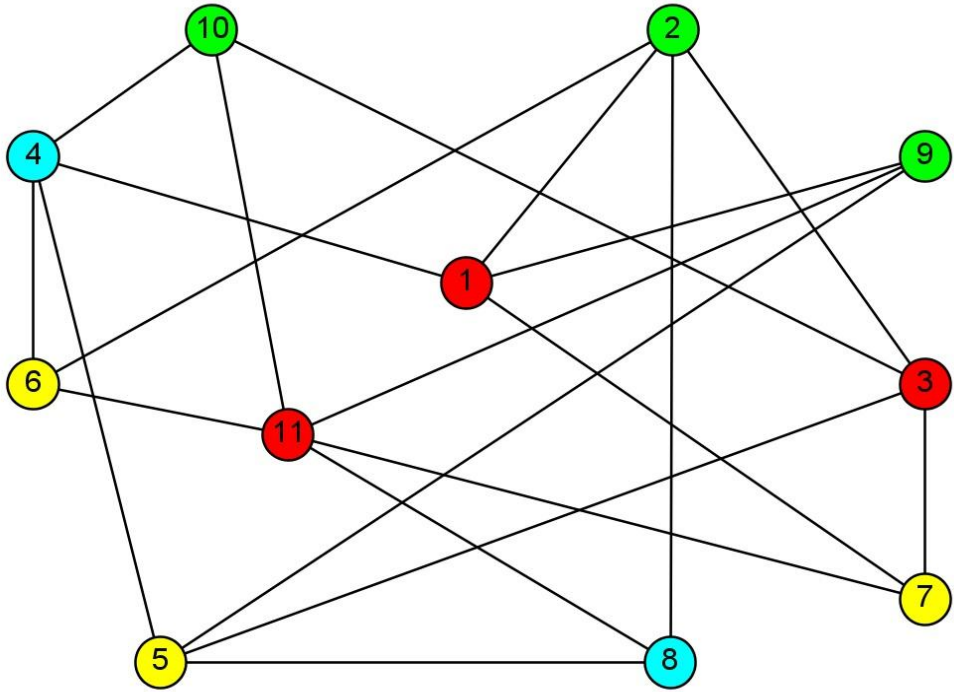

Figure 4.4: Valid chromosome with minimum chromatic colour

In this process we reduce chromatic number 6 to 4 and again running this process does not reduce any colour. This is minimum expected colour of standard DIMACS benchmark data sets.

## 4.11 Generation

In this process of evolutionary algorithm generation or iteration is one of the main terms of this process. When iteration is finish parent chromosomes are replaced by the offspring's which is best comparing to their parents. This process will go on until the best chromosome will not found. When it is found, this process will terminate by itself. This is how evolutionary algorithm works.

Here we give some example of minimum colour of [4] and our obtained results are bellow

| Data File($G$) | Node | Edge | $\chi(G)$ | **EAGCP** | [1] | [3] | [25] |
|---|---|---|---|---|---|---|---|
| myciel3.col | 11 | 20 | 4 | 4 | 4 | 4 | 4 |
| myciel4.col | 23 | 71 | 5 | 5 | 5 | 5 | 5 |
| queen5_5.col | 25 | 160 | 5 | 5 | 5 | 5 | 5 |
| queen6_6.col | 36 | 290 | 7 | 8 | 7 | 7 | 7 |
| myciel5.col | 47 | 236 | 6 | 6 | 6 | 6 | 6 |
| huck.col | 74 | 301 | 11 | 11 | 11 | 11 | 11 |
| jean.col | 80 | 254 | 10 | 10 | 10 | 10 | 10 |
| anna.col | 138 | 493 | 11 | 12 | 11 | 11 | 11 |
| david.col | 87 | 406 | 11 | 12 | 11 | 11 | 11 |

Table 4.17: Obtained result for data file in DIMACS [4] with others

Here $\chi(G)$ represent minimum expected number of different dataset and comparing that obtained result to the recent paper which is found in reference.

# Chapter 5
# Experimental Result

We use difference data set for DIMACS

For example

## 5.1 Output Result of myciel3.col

Node: 11

Edge: 20

Generation: 51

Chi (G): 4

EAGCP: 4

Population Size: 50

Maximum Colour: 6

Mutation Probability: 10%

Additional Probability: 10%

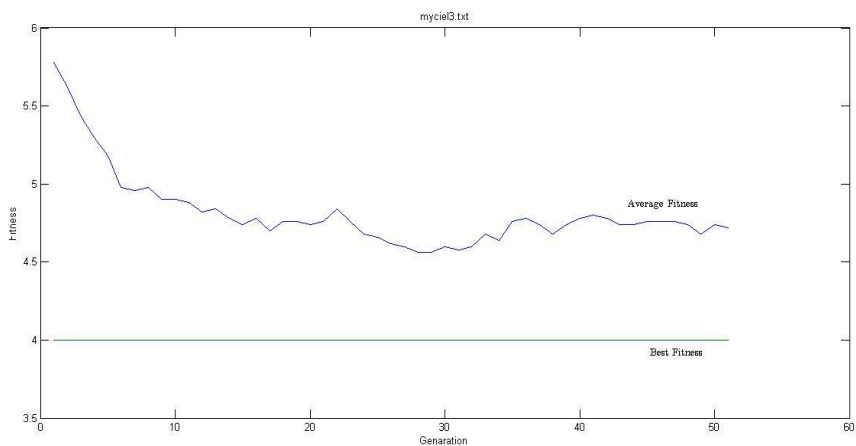

Figure 5.1: Output Graph of myciel3.col

## 5.2 Output Result of myciel4.col

Node: 23

Edge: 71

Generation: 163

Chi (G): 5

EAGCP: 5

Population Size: 50

Maximum Colour: 12

Mutation Probability: 10%

Additional Probability: 10%

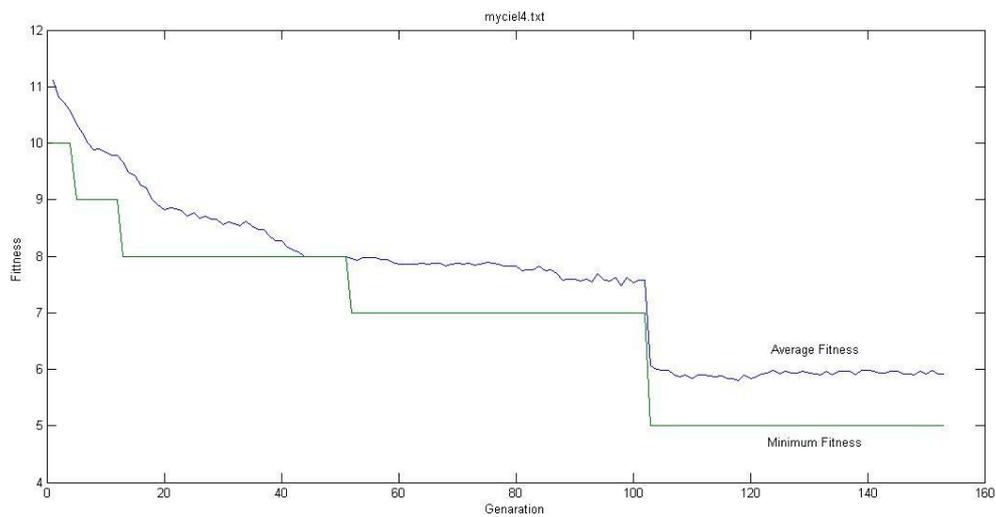

Figure 5.2: Output Graph of myciel4.col

## 5.3 Output Result of myciel5.col

Node: 47

Edge: 236

Generation: 765

Chi (G): 6

EAGCP: 6

Population Size: 150

Maximum Colour: 24

Mutation Probability: 10%

Additional Probability: 10%

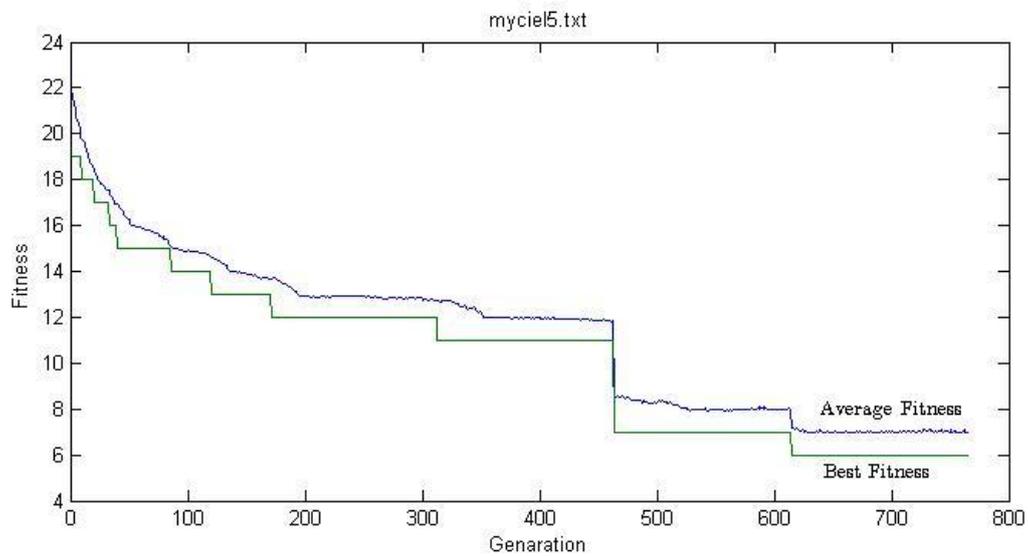

Figure 5.3: Output Graph of myciel5.col

## 5.4 Output Result of queen5_5

Node: 25

Edge: 160

Generation: 845

Chi (G): 5

EAGCP: 5

Population Size: 200

Maximum Colour: 17

Mutation Probability: 15%

Additional Probability: 10%

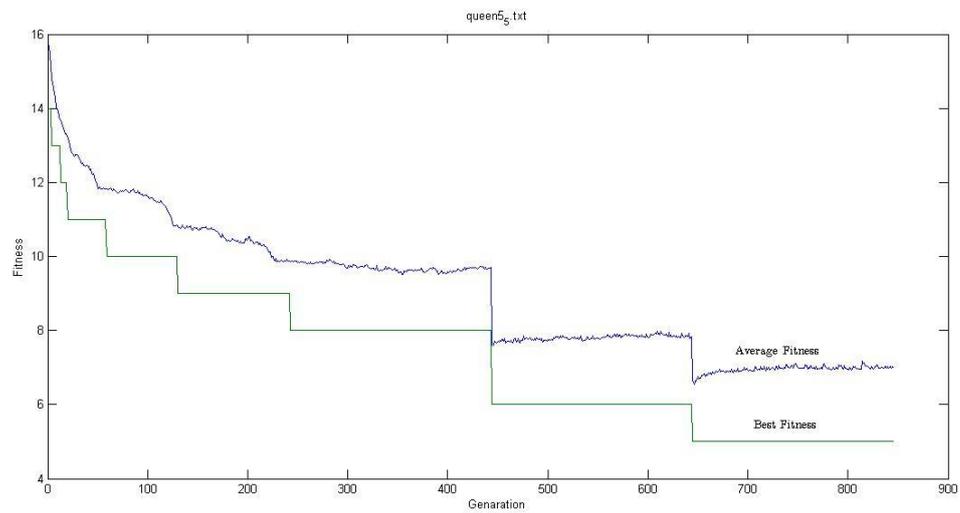

Figure 5.4: Output Graph of queen5_5

## 5.5 Output Result of huck.col

Node: 71

Edge: 301

Generation: 1897

Chi (G): 11

EAGCP: 11

Population Size: 500

Maximum Colour:  54

Mutation Probability: 10%

Additional Probability: 10%

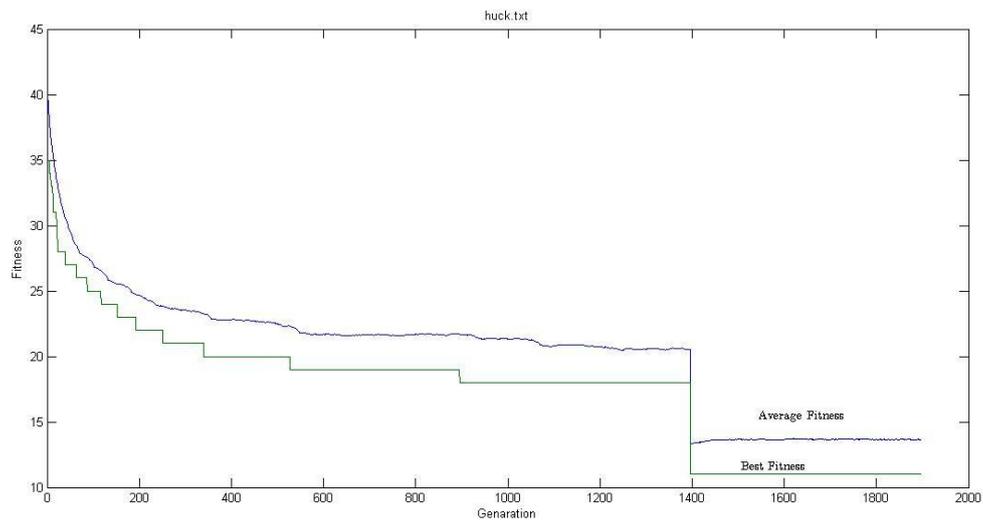

Figure 5.5: Output Graph of huck

## 5.6 Output Result of david.col

Node: 87

Edge: 406

Generation: 1937

Chi (G): 11

EAGCP: 12

Population Size: 500

Maximum Colour: 83

Mutation Probability: 10%

Additional Probability: 10%

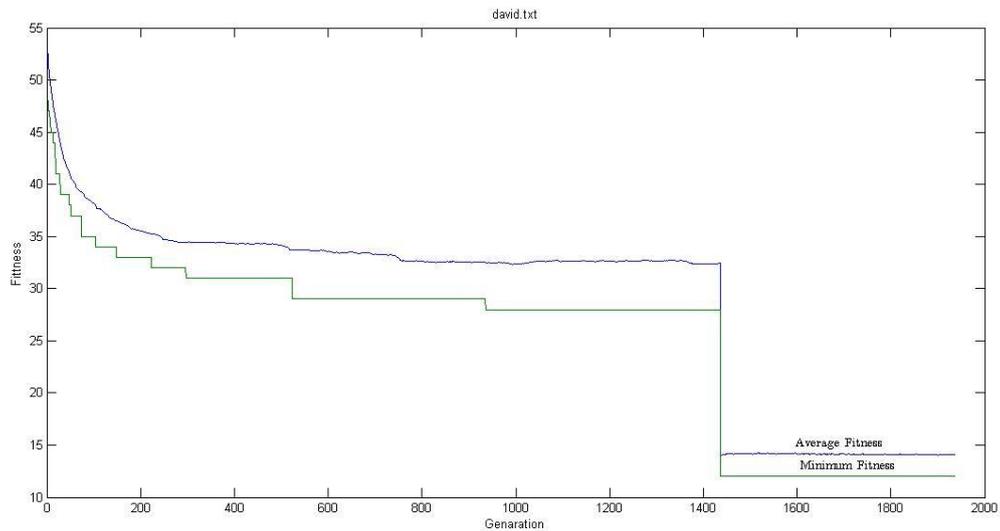

Figure 5.6: Output Graph of david

# Chapter 6
# Conclusion and Future work

In this thesis we propose Evolutionary Algorithm for GCP. In this paper we use binary encoding to reduce colour of GCP.

"Of course there is no best possible way to sort; we must define precisely what is meant by "best", and there is no best possible way to define "best"[Donald E. Knuth, in the context of sorting] [16]."

We have tried our algorithm is different data set and the output gives a near optimal in some problem and some are optimal. Increasing the number of Total population and changing the mutation probability will help us to find all the data sets optimal solution more preciously and efficiently. In this research we try to solve Discrete Mathematics and Theoretical Computer Science graph. Many graphs are given expected solution but for some of the large graphs our result is higher than the optimal. In future we will try to solve those large graphs efficiently and minimize the time complexity for every data set.

In future we will try to tested large data file of standard DIMACS benchmark and try to find expected result or less than present result.

# Reference

[01]  Musa M. Hindi and Roman V. Yampolskiy, "Genetic Algorithm Applied to the Graph Colouring Problem", In *proc.23rd Midwest Artificial Intelligence and Cognitive Science Conference* 2012 Cincinnati, USA, Apr 21-22, 2012

[02]  Martin Schwengerer, "Algorithm Selection for the Graph Colouring Problem", MASTER'S THESIS, *The Faculty of Informatics at the Vienna University of Technology Vienna*, 18.10.2012

[03]  Biman Ray,Anindya J Pal,Debnath Bhattacharyya,Tai-hoon Kim, "An Efficient GA with Multipoint Guided Mutation for Graph Colouring Problems", *International Journal of Signal Processing, Image Processing and Pattern Recognition* Vol. 3, No. 2, June 2010

[04]  M. Trick, "Graph Colouring Instances" (Michael Trick's Operations Research Page) [online], http://mat.gsia.cmu.edu/COLOUR/instances.html (Accessed: 24 February 2013).

[05]  Gwee, B. H., Lim, M. H., and Ho, J. S,"Solving four- colouring map problem using genetic algorithm", *First New Zealand International Two-Stream Conference on Artificial Neural Networks and Expert Systems*, 332-333. New Zealand

[06]  Marx, Daniel, and Marx, D Aniel. 2004, "Graph Colouring Problems and Their Applications in Scheduling", *John von Neumann PhD Students Conference*,Budapest, Hungary O'Madadhain, Joshua, Fisher, Danyel, and Nelson, Tom. 2011.

[07]  Hale, W. K. 1980,"Frequency assignment: Theory and applications", *Proceedings of the IEEE* 12: 1497-1514

[08]  Shengning, Wu, and Sikun, Li. 2007,"Extending Traditional Graph-Colouring Register Allocation Exploiting Meta-heuristics for Embedded Systems" In Proceedings of *The Third International Conference on Natural Computation*. ICNC, 324-329. Haikou, Hainan, China

[09]  Surowiecki, J.,."The Wisdom of Crowds.GardenCity:Doubleday",*Introduction, Kapitel 1 & 2* (pp. XI-XXI & 3-39)

[10]  Edward Vul and Harold Pashler, "Measuring the Crowd within Probabilistic Representations within Individuals",*PSYCHOLOGICAL SCIENCE,*RECEIVED 9/17/07; REVISION ACCEPTED 1/7/08

[11]  Sheng Kung Michael Yi, Mark Steyvers, Michael D. Lee,Matthew J. Dry, "wisdom of the crowds in travelling salesman problem"

[12]  David L. Applegate,Robert E. Bixby b VaˇsekChv´atal,William Cook, Daniel G. Espinoza,MarcosGoycoolea,KeldHelsgaun, "CertificationofanoptimalTSPtourthrough85,900cities", 2006

[13] Sarah Carruthers, Michael E. J. Masson and Ulrike Stege, "Human Performance on Hard Non-Euclidean Graph Problems: Vertex Cover", *the Journal of Problem Solving:* Vol. 5: Iss-1, Article 5, Approximation algorithms. Berlin: Springer-Verlag.

[14] IRIS VAN ROOIJ, ULRIKE STEGE, andALISSA SCHACTMAN, "Convex hull and tour crossings in the Euclidean traveling salesperson problem: Implications for human performance studies", *Memory & Cognition* 2003, 31 (2), 215-220.

[15] Alice E. Smith and David W. Coit, "Constraint-Handling Techniques - Penalty Functions," *Handbook of Evolutionary Computation, Institute of Physics Publishing and Oxford University Press*, Bristol, U.K., 1997, Chapter C5.2.

[16] D. E. Knuth,"The art of computer programming",*Number Bd. 3 in Addison-Wesley series in computer science and information processing*, Addison-Wesley, 1981.

[17] SANCHITA PAUL,ANIRBAN DE SARKAR, "SOLUTION TO THE 0/1 KNAPSACK PROBLEM BASED ON DNA COMPUTING", *Journal of Theoretical and Applied Information Technology* 2005 - 2008 JATIT.

[18] Xiao-FengXie, Jiming Liu, "Graph Colouring by Multi agent Fusion Search", *Journal of Combinatorial Optimization*, 2009. 18(2): 99-123

[19] Greg Durrett, Muriel Médard and Una-May O'Reilly, "A Genetic Algorithm to Minimize Chromatic Entropy", *P. Cowling and P. Merz (Eds.): EvoCOP 2010, LNCS 6022,* pp. 59–70, 2010.Springer-Verlag Berlin Heidelberg 2010

[20] Kubale, M. (2004), Graph Colourings, American Mathematical Society, ISBN 0-8218-3458-4

[21] Van Lint, J. H.; Wilson, R. M. (2001),"A Course in Combinatorics (2nd ed.)", *Cambridge University Press*, ISBN 0-521-80340-3.

[22] Jensen, T. R.; Toft, B. (1995), "Graph Colouring Problems,"*Wiley-Interscience, New York,* ISBN 0-471-02865-7

[23] D.E. Glodberg, "Genetic Algorithm in Search, Optimization & Machine Learning", Published by Pearson Education (Singapore) Pte Ltd, ISBN 81-7808-130-X

[24] S. Rajasekaran, G. A. VijayalakshmiPai, "Neural Networks, Fuzzy Logic and Genetic Algorithm: Synthesis and Applications", ISBN-978-81-203-2186-1

[25] Abbasian, Reza, and MalekMouhoub, "An efficient hierarchical parallel genetic algorithm for graph coloring problem",*Proceedings of the 13th Annual Genetic and Evolutionary Computation Conference. 2011.*